\documentclass[letterpaper, 10 pt, conference]{ieeeconf}  % Comment this line out if you need a4paper

\IEEEoverridecommandlockouts                              % This command is only needed if 
\def\etal{\emph{et al.}}

\usepackage{amsfonts}
\usepackage{amsmath}
\usepackage{amssymb}

\usepackage{fixltx2e} % for subscripts

\usepackage{hyperref}
\usepackage{cite}
\usepackage[pdftex]{graphicx}

\title{\LARGE \bf
Fast Scene Understanding for Autonomous Driving
}

\author{Davy Neven, Bert De Brabandere, Stamatios Georgoulis, Marc Proesmans and Luc Van Gool\\
ESAT-PSI, KU Leuven\\
{\tt\small firstname.lastname@esat.kuleuven.be}
}% <-this % stops a space

\begin{document}

\maketitle
\thispagestyle{empty}
\pagestyle{empty}

%%%%%%%%%%%%%%%%%%%%%%%%%%%%%%%%%%%%%%%%%%%%%%%%%%%%%%%%%%%%%%%%%%%%%%%%%%%%%%%%

\begin{abstract}

Most approaches for instance-aware semantic labeling traditionally focus on accuracy. Other aspects like runtime and memory footprint are arguably as important for real-time applications such as autonomous driving.  
Motivated by this observation and inspired by recent works that tackle multiple tasks with a single integrated architecture~\cite{uhrig2016pixel, teichmann2016multinet, kokkinos2016ubernet}, in this paper we present a real-time efficient implementation based on ENet~\cite{paszke2016enet} that solves three autonomous driving related tasks at once: semantic scene segmentation, instance segmentation and monocular depth estimation.
Our approach builds upon a branched ENet architecture with a shared encoder but different decoder branches for each of the three tasks.
The presented method can run at 21 fps at a resolution of 1024x512 on the Cityscapes dataset without sacrificing accuracy compared to running each task separately.

\end{abstract}

%%%%%%%%%%%%%%%%%%%%%%%%%%%%%%%%%%%%%%%%%%%%%%%%%%%%%%%%%%%%%%%%%%%%%%%%%%%%%%%%

\section{INTRODUCTION}

The last years the re-appearance of Convolutional Neural Networks (CNNs), whose origin traces back to the 1970s and 1980s, has led to significant advances in many computer vision tasks, such as image classification~\cite{krizhevsky2012imagenet}, object detection~\cite{girshick2014rich}, semantic scene segmentation~\cite{long2015fully}, instance segmentation~\cite{hariharan2014simultaneous}, and monocular depth estimation~\cite{eigen2014depth} to name a few.
The majority of these works rely on fine-tuning or slightly altering a CNN architecture, typically the VGG network~\cite{simonyan2014very}, resulting in task-specific CNNs with long inference times that each require a single GPU to run.
Admittedly, this is not enough for autonomous driving applications where many of the aforementioned tasks should run in parallel, in real-time, and on a limited number of GPU devices.
Furthermore, as shown in recent works~\cite{uhrig2016pixel, teichmann2016multinet, kokkinos2016ubernet} there is merit in combining multiple tasks in a single integrated architecture, as one task might benefit from another leaving smaller space for 'blindspots', which is crucial for self-driving vehicles.

\begin{figure}[t]
	\begin{center}
		\includegraphics[width=1\linewidth]{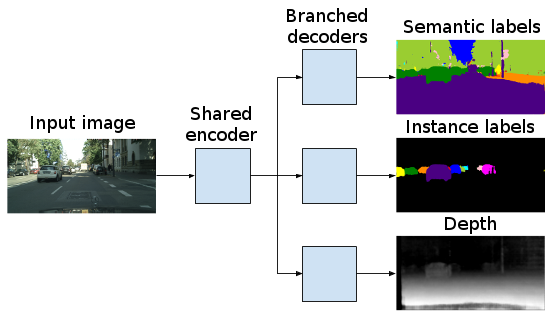}
	\end{center}
	\caption{Overview of our method. From left to right: the input image is passed though the encoding step of an ENet-inspired architecture to create feature maps, which in turn are forwarded to different branches that perform decoding to arrive at the three outputs, i.e. semantic labels, instance labels and depth. A video is available at \url{https://youtu.be/55ElRh-g_7o}.}
	\label{fig:method_overview}
\end{figure}

Motivated by these observations, in this paper we focus on street scene understanding and present an efficient implementation that combines the tasks of semantic scene segmentation, instance segmentation, and monocular depth estimation.
Unlike state-of-the-art methods, that use networks with huge number of parameters and long inference times (e.g. VGG~\cite{simonyan2014very}, SegNet~\cite{badrinarayanan2015segnet}, FCN~\cite{long2015fully}), we build upon a real-time architecture, in particular ENet~\cite{paszke2016enet} that has proven to offer image processing rates higher than 10 fps on a single GPU device.
Specifically, we use a common ENet encoding step for all tasks, but introduce a branched ENet architecture for the decoding step (i.e. one branch for each of the three different tasks).
Fig.~\ref{fig:method_overview} gives an overview of our approach.

Although we do not introduce a new architecture, in this paper we show how to efficiently combine existing components to build a solid architecture for real-time scene understanding.
In Sec.~\ref{sec:related_work} we describe related work on integrated architectures that tackle multiple tasks.
Next, we present the implementation details of our method in Sec.~\ref{sec:method}.
Finally, in Sec.~\ref{sec:results} and~\ref{sec:conclusion} we report results for each of the tasks and provide some insights into the strengths and limitations of the presented approach.

%%%%%%%%%%%%%%%%%%%%%%%%%%%%%%%%%%%%%%%%%%%%%%%%%%%%%%%%%%%%%%%%%%%%%%%%%%%%%%%%

\section{RELATED WORK}
\label{sec:related_work}

The amount of research performed in literature on the three main tasks studied in this paper, i.e. semantic scene segmentation, instance segmentation, and monocular depth estimation, is vast.
In what follows, we solely focus on related works that have combined one or more of these tasks in a single integrated architecture.

Eigen and Fergus~\cite{eigen2015predicting} addressed the tasks of depth prediction, surface normal estimation, and semantic labeling using a multiscale convolutional network architecture that progressively refines predictions from a sequence of scales.
Uhrig~\etal~\cite{uhrig2016pixel} presented a method that leverages a FCN network to predict semantic labels, depth, and an instance-based encoding using each pixel's direction towards its corresponding instance center and consequently applying low-level computer vision techniques.
Kokkinos~\cite{kokkinos2016ubernet} went one step further from the previous approaches, and introduced a CNN, namely UberNet, that jointly handles low-, mid-, and high-level vision tasks in a unified architecture.
His universal network tackles boundary detection, normal estimation, saliency estimation, semantic segmentation, human part segmentation, semantic boundary detection, region proposal generation, and object detection.
Despite obtaining competitive performance while jointly addressing many different tasks, all these approaches suffer from poor inference times making them unsuitable for real-time autonomous driving applications with high frame-rate demands.

Recently, Teichmann~\etal~\cite{teichmann2016multinet} argued that improving the computational times is more important than improving performance, especially for the case of self-driving vehicles.
They presented an approach to joint classification, detection, and semantic segmentation via a unified architecture where the encoder is shared amongst the three tasks, marginally reaching a computational time of 10 fps on the KITTI dataset.
Our approach also focuses on further improving the computational times but addresses different tasks, in particular semantic scene segmentation, instance segmentation, and monocular depth estimation, and achieves a computational time of 21 fps on the Cityscapes dataset. 
To our knowledge this is the first system to estimate depth, semantic and instance segmentation at these frame-rates. 

%%%%%%%%%%%%%%%%%%%%%%%%%%%%%%%%%%%%%%%%%%%%%%%%%%%%%%%%%%%%%%%%%%%%%%%%%%%%%%%%

\section{METHOD}
\label{sec:method}

In order to predict depth, semantic and instance segmentation in real-time, we modify the ENet architecture into a multi-branched network, having three output branches, one for each task (see Fig.~\ref{fig:method_overview}).
The original network, as described in~\cite{paszke2016enet}, consists of an encoding step that has three stages (stage 1, 2, 3) and a decoding step that has two stages (stage 4, 5).
Since the ENet decoding step is merely for upscaling and finetuning the output of the encoding step, sharing the full encoder (stages 1, 2, 3) between all branches would lead to poor results.  
Instead, our multi-branch network is constructed as follows: our shared "encoder" consists of stages 1 and 2 of the original Enet network, before continuing to each branch that combines stage 3 of the original ENet encoder with stages 4 and 5 of the original ENet decoder. 
In what follows, we dive into the details of the individual branches that each performs one task.

\textbf{Semantic segmentation}
The semantic segmentation branch is trained using the standard pixel-wise cross-entropy loss. % MAYBE CITATION?
The classes are weighted using the method described in~\cite{paszke2016enet} and trained until convergence. 
The semantic segmentation is used for free space detection as well for classifying the objects found by the instance segmentation branch. 

\textbf{Instance segmentation}
In order to perform instance segmentation using a typical feed-forward network without having to resort to slower detect-and-segment approaches, we use a recently introduced discriminative loss function~\cite{discriminativeInst} suited for real-time instance segmentation that can be plugged into an off-the-shelf network.
The intuition behind the proposed loss function is that pixel embeddings (i.e. the network's output for each pixel) with the same label (i.e. same instance) should end up close together, while embeddings with a different label (i.e. different instance) should end up far apart.

Inspired by Weinberger~\etal~\cite{weinberger2009distance} and other distance metric learning approaches, 
the authors propose a loss function with two competing terms to achieve this objective: a \textit{variance term} pulling pixel embeddings towards the mean embedding of their cluster, and a \textit{distance term} pushing the clusters away from each other. 
To relax the constraints on the network, the variance and distance terms are hinged: embeddings within a distance of $\delta_v$ from their cluster centers are no longer attracted to it and cluster centers further apart than $2\delta_v$ are no longer repulsed. A small regularization pull-force that draws all clusters towards the origin keeps the activations bounded. 
These three terms can be written as follows, with $C$ the number of clusters in ground truth, $N_c$ the number of elements in cluster $c$, $x_i$ an embedding, $\mu_c$ the mean embedding of cluster $c$, $\lVert \cdot \rVert$ the L2 distance, and $\left[ x \right]_{+} = \textrm{max}(0,x)$ denotes the hinge: 

\begin{equation}
\begin{cases}
L_{var} = \frac{1}{C} \sum_{c=1}^{C} \frac{1}{N_c} \sum_{i=1}^{N_c} \left[ \lVert \mu_c - x_i \rVert - \delta_{\textrm{v}} \right]_{+}^2 \\ \\
L_{dist} = \frac{1}{C (C-1)} \mathop{\sum_{c_A = 1}^{C} \sum_{c_B = 1,}^{C}}_{c_A \neq c_B} \left[ 2 \delta_{\textrm{d}} - \lVert \mu_{c_A} - \mu_{c_B} \rVert \right]_{+}^2 \\ \\
L_{reg} = \frac{1}{C} \sum_{c=1}^{C} \lVert \mu_{c} \rVert 
\end{cases}
\end{equation}

The final loss can then be written as the sum of the above terms: $L_{inst} = L_{var} + L_{dist} + L_{reg}$. 
When the loss has converged, all pixel embeddings are within a distance of $\delta_v$ from their cluster center and all cluster centers are at least $2 \delta_d$ apart. 
By setting $\delta_d > 2 \delta_v$, each embedding is closer to all embeddings of its \textit{own} cluster than to any embedding of a \textit{different} cluster. 
During inference we can then threshold with bandwidth $b = \delta_v$ around any embedding to select all embeddings belonging to the same cluster. 
Since the loss on the test set will not be zero, we apply a GPU accelerated variant of the mean-shift algorithm~\cite{fukunaga1975estimation} to shift to a center pixel around which we threshold, avoiding outliers.

\textbf{Depth estimation from a single image}
The standard loss used in most regression problems, like monocular depth estimation, is the $L2$ loss. It minimizes the difference between predicted $D$ and ground truth $D^{*}$ depth: $L2(D,D^{*})~=~\frac{1}{n}\sum_{i}d_{i}^{2}$, with $d~=~D-D^{*}$.
Recently, Eigen and Fergus~\cite{eigen2015predicting} added two more terms to the typical $L2$ loss for the depth estimation task; one for scale invariance ($-\frac{1}{2n^{2}}(\sum_{i}d_{i})^{2}$), and another for similarity in local structure ($\frac{1}{n}\sum_{i}[(\nabla_{x}d_{i})^{2}+(\nabla_{y}d_{i})^{2}]$, with $\nabla_{x}$ and $\nabla_{y}$ denoting the horizontal and vertical image gradients).
Instead, the depth estimation branch uses the reverse Huber loss (berHu)~\cite{owen2007robust},
\begin{equation}
L_{dep}~=~
\begin{cases}
|d| & |d| \le c \\
\frac{d^{2}+c^{2}}{2c} & |d| > c,
\end{cases}
\end{equation}
that shows a good balance between penalizing high residuals that usually account for the mean depth and low residuals that explain the smaller depth details.
We have experimentally found that this choice yields a better final error than using the $L2$ loss, even with the added terms.
Notice that, the reverse Huber loss formulation above is continuous and first order differentiable at point $c$, which is set to $c~=~\frac{1}{5}\text{max}_{i}(d_{i})$ as in~\cite{laina2016deeper}. 
We use the SGM-calculated disparity depth maps of the Cityscapes dataset as ground truth for this task. 

\textbf{Training}
To train our multi-task network, the three losses described above are summed and equally weighted.
Although different weights can also be used for each task we found that using equal weights already leads to good performance.
We start from a pretrained encoder, trained for Cityscapes segmentation, and continue training the three tasks together. 
We train with a batch size of 10 at a resolution of 1024x512 and use Adam with a learning rate of 5e-4.
Note that, we keep the parameters of the batch norm layers fixed.

%%%%%%%%%%%%%%%%%%%%%%%%%%%%%%%%%%%%%%%%%%%%%%%%%%%%%%%%%%%%%%%%%%%%%%%%%%%%%%%%

\section{RESULTS}
\label{sec:results}
\textbf{Semantic and instance segmentation}
\begin{table}
    \caption{Semantic segmentation results on the Cityscapes benchmark.}
	\begin{center}
		\begin{tabular}{l|c|c}
			& IoU class & IoU category \\
			\hline
            Segnet~\cite{badrinarayanan2015segnet} & $56.1$ & $79.8$ \\
            ENet~\cite{paszke2016enet} & $58.3$ & $80.4$ \\
            SQ~\cite{tremlspeeding} & $59.8$ & $84.3$ \\
            \hline
			Ours & $59.3$ & $80.4$ \\
		\end{tabular}
	\end{center}
	\label{tab:resultsSemseg}
\end{table}
\begin{table}
    \caption{Instance segmentation results on the Cityscapes benchmark.}
	\begin{center}
		\begin{tabular}{l|c|c|c|c}
			& AP & AP0.5 & AP100m & AP50m \\
			\hline
            InstanceCut~\cite{kirillov2016instancecut} & $23.7$ & $44.8$ & $38.9$ & $42.5$ \\
            PPLoss & $24.4$ & $43.2$ & $40.0$ & $44$ \\
            Pixelwise DIN~\cite{arnab2017pixelwise} & $25.7$ & $45.7$ & $41.1$ & $44.2$ \\
            DWT~\cite{bai2016deep} & $31.5$ & $48.5$ & $50.8$ & $53.5$ \\
            Shape-aware~\cite{hayder2016shape} & $35.7$ & $54.7$ & $58.2$ & $63.1$ \\
            SGN & $39.4$ & $59.7$ & $60.1$ & $63.2$ \\
            Mask R-CNN~\cite{he2017mask} & $46.9$ & $68.3$ & $65.5$ & $67.4$\\
            \hline
			Ours & $21.0$ & $38.6$ & $34.8$ & $38.7$ \\
		\end{tabular}
	\end{center}
	\label{tab:resultsInstseg}
    \vspace{-1.0em}
\end{table}
We report Cityscapes semantic segmentation results in Tab.~\ref{tab:resultsSemseg} and instance segmentation results on the car class in Tab.~\ref{tab:resultsInstseg}. 
We notice that by jointly training our network for 3 different tasks, we match and even slightly outperform standard ENet for semantic scene segmentation. 
This justifies our hypothesis that training with multiple tasks at once can increase the performance of each individual task.

As expected, our result for instance segmentation lacks behind the other methods on the Cityscapes benchmark, since they are all optimized for accuracy and are far from real-time. 
They either rely on a big network or use highly accurate pre-generated semantic segmentation labels, which explains their significantly higher performance, compared to our result. 
Nevertheless, this work can serve as a baseline for methods that also focus on speed. 

\begin{figure}[t]
	\begin{center}
		\includegraphics[width=1\linewidth]{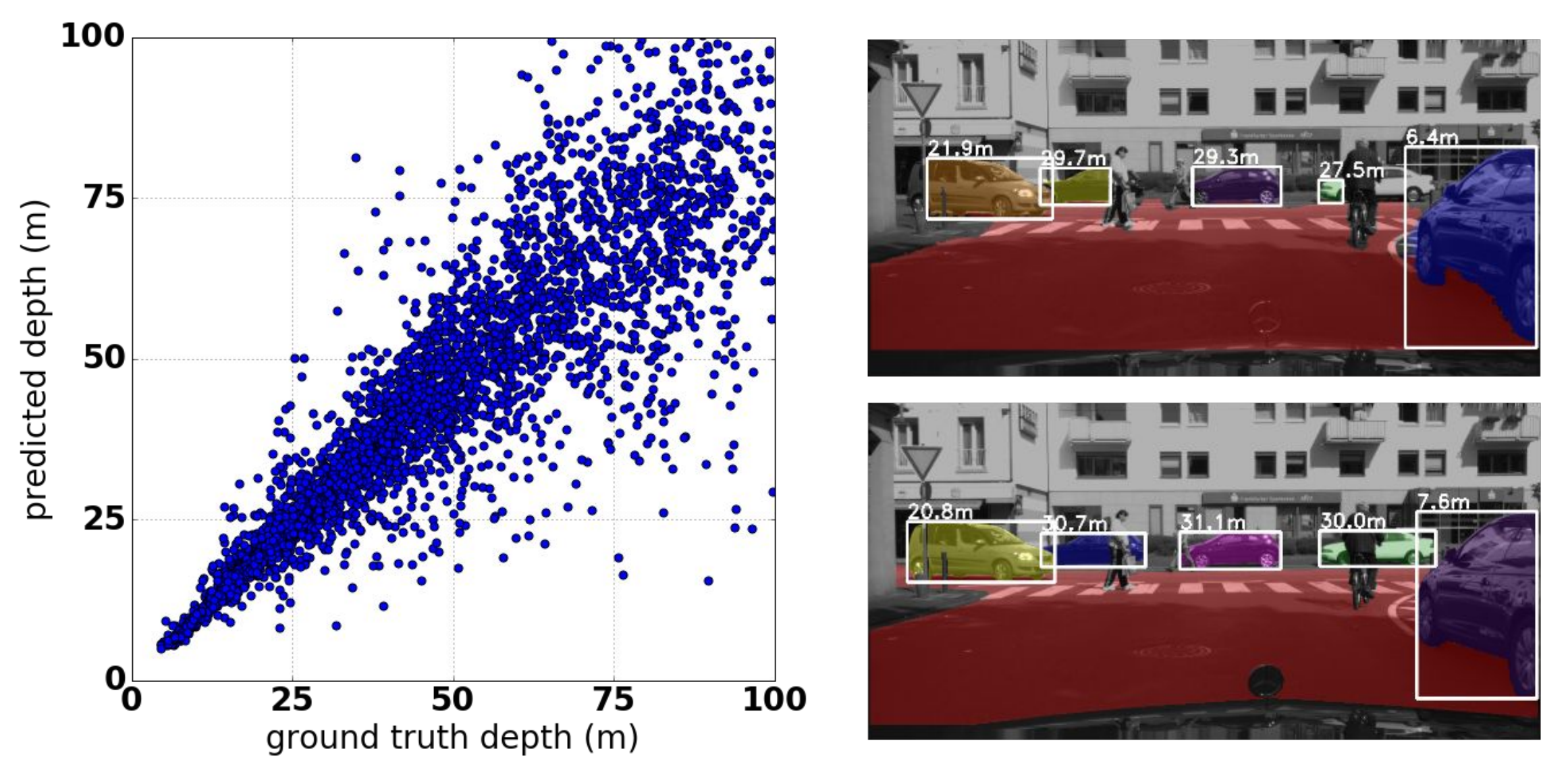}
	\end{center}
	\caption{Left: ground truth and predicted depth for each car in the dataset. The expected trend of nearby cars being predicted more accurately than far-away cars is clearly visible. Right: a qualitative result (top: prediction, bottom: ground truth).}
	\label{fig:depth_confusion}
\end{figure}

\textbf{Depth}
In Fig.~\ref{fig:depth_confusion} we plot for each car in the dataset its ground truth depth versus its predicted depth, which is calculated as an average over the predicted depth map masked out with the ground truth instance mask. 
The expected trend of nearby cars being predicted more accurately than far-away cars is clearly visible. 
Some of the extreme outliers are caused by cars that are mostly occluded and thus only consist of a few pixels.
These extreme cases can in principle be detected and filtered out using the instance mask. 
We encourage others to include similar plots in their work on car depth estimation, as it is more informative than a single summary number.

Nevertheless, we follow~\cite{uhrig2016pixel} and report three metrics in Tab.~\ref{tab:resultsDepth}: mean absolute error (MAE), root mean squared error (RMSE) and absolute relative error (ARD). 
Note that we calculate the depth of each car by average pooling the predicted depth map with the \textit{ground truth} instance masks.
This is unlike~\cite{uhrig2016pixel}, who calculate the depth with the \textit{predicted} instance masks, and report the metrics only over predicted cars that match with ground truth cars. 
This means that the metric they report does not take the depth estimation of undetected smaller or badly visible cars into account, leading to a number that is dependent on the instance segmentation performance. 
By reporting the numbers over the \textit{ground truth} car masks we avoid this entanglement, but some caution is necessary when comparing the numbers. We provide the numbers at different maximum depths of 100m, 50m and 25m.
\begin{figure*}[t]
	\begin{center}
		% \fbox{\rule{0pt}{2in} \rule{.9\linewidth}{0pt}}
		\includegraphics[width=1.0\linewidth]{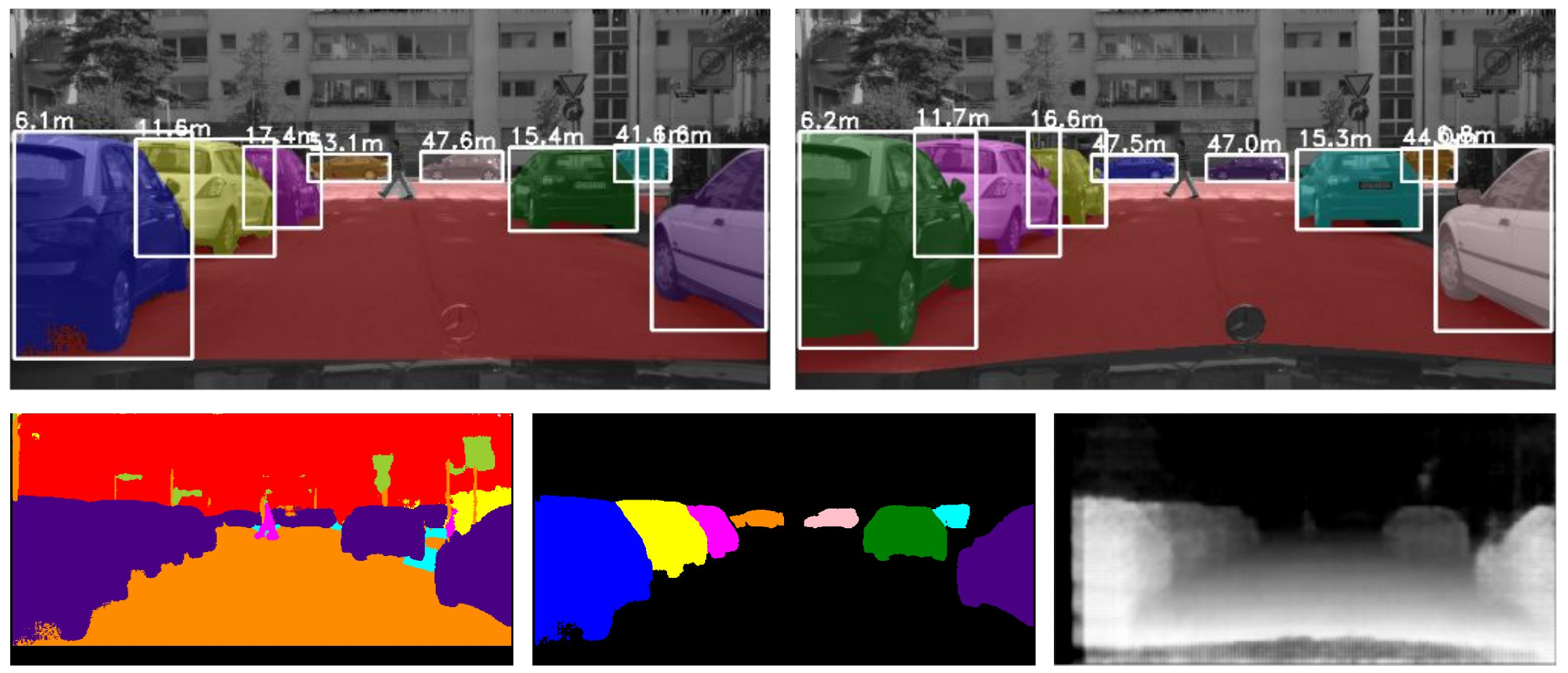}
	\end{center}
	\caption{Another qualitative result on the validation set. Top: predicted and ground truth summary picture. Bottom: predicted semantic, instance and depth maps.}
	\label{fig:method_results2}
\end{figure*}
\begin{table}
    \caption{Single image depth results on the Cityscapes validation set. Some caution necessary when comparing with~\cite{uhrig2016pixel} (sect.~\ref{sec:results}).}
	\begin{center}
		\begin{tabular}{l|c|c|c}
			& MAE & RMSE & ARD \\
			\hline
			Uhrig~\etal~\cite{uhrig2016pixel} (val) & $7.7$m & $24.8$m & $11.3$\%\\
            \hline
            Ours (val, $<$ 100m) & $7.5$m & $11.9$m & $13.8$\%\\
            Ours (val, $<$ 50m) & $3.5$m & $5.6$m & $11.0$\%\\
            Ours (val, $<$ 25m) & $1.5$m & $2.5$m & $8.9$\%\\
		\end{tabular}
	\end{center}
	\label{tab:resultsDepth}
    \vspace{-1.0em}
\end{table}

\textbf{Multi-task network and speed}
In Tab.~\ref{tab:ablationStudy} we provide a comparison between training the tasks separately (each running on an ENet of their own), versus training them together with a shared encoder as explained in the previous section. The benefits of training the three tasks together in a single multi-task network are clear: the speed almost doubles and the memory usage decreases drastically. 
This makes our approach suitable for real-time autonomous driving applications that require a low memory footprint. 
Important to note is that the accuracy of the individual tasks does not decrease when training together: in fact we even notice a slight performance increase. 
This suggests that the shared encoder can effectively learn to exploit the common structure of the three related semantic tasks.

\begin{table}
    \caption{Semantic segmentation (IoU\textsubscript{c}), instance segmentation (AP), depth (MAE\textsubscript{$100$m} on the val set), memory, and speed (forward pass on a Pascal TitanX) at test time when trained separately versus together.}
	\begin{center}
		\begin{tabular}{l|c|c|c|c|c}
			& semantic & instance & depth & mem & speed \\
			\hline
            Trained separately & $58.3$\% & $0.20$\% & $9.2$m & $2.6$ GB & $12$ fps \\
            Trained together & $59.3$\% & $0.21$\% & $7.5$m & $1.2$ GB & $21$ fps \\
		\end{tabular}
	\end{center}
	\label{tab:ablationStudy}
    \vspace{-1.5em}
\end{table}

%%%%%%%%%%%%%%%%%%%%%%%%%%%%%%%%%%%%%%%%%%%%%%%%%%%%%%%%%%%%%%%%%%%%%%%%%%%%%%%%

\section{CONCLUSION}
\label{sec:conclusion}
Overall, our system is fast but lags behind the state-of-art in terms of segmentation accuracy. 
Nevertheless, we believe that it can serve as a low-complexity baseline for other multi-task approaches that focus on speed, and as a starting point for further exploration of the speed-accuracy trade-off in scene understanding.
Furthermore, we observe that jointly training tasks can potentially lead to increased performance.

\noindent {\bf Acknowledgement:} The work was supported by Toyota, and was carried out at the TRACE Lab at KU Leuven (Toyota Research on Automated Cars in Europe - Leuven).

%%%%%%%%%%%%%%%%%%%%%%%%%%%%%%%%%%%%%%%%%%%%%%%%%%%%%%%%%%%%%%%%%%%%%%%%%%%%%%%%

\bibliographystyle{plain}
\bibliography{./ref}

\end{document}